\documentclass[10pt,twocolumn,twoside]{IEEEtran}

\usepackage{cite}
\usepackage{graphicx}
\usepackage[cmex10]{amsmath}
\usepackage{mdwmath}
\usepackage[tight,footnotesize]{subfigure}
\usepackage{url}
\begin{document}

\title{Performing Nonlinear Blind Source Separation \\ with Signal Invariants}

\author{David~N.~Levin%
\thanks{D. Levin is with the Department of Radiology and the Committee on Medical Physics, University of Chicago. Mailing address: 1720 N. Lasalle Dr., Unit 25, Chicago, IL 60614. Telephone: 312-482-8624. Fax: 312-482-9758. Email: d-levin@uchicago.edu}%
\thanks{Manuscript received April 1, 2009}}

\maketitle

\begin{abstract}
Given a time series of multicomponent measurements $x(t)$, the usual objective of nonlinear blind source separation (BSS) is to find a ``source" time series $s(t)$, comprised of statistically independent combinations of the measured components. In this paper, the source time series is required to have a density function in $(s,\dot{s}) \mbox{-space}$ that is equal to the product of density functions of individual components. This formulation of the BSS problem has a solution that is unique, up to permutations and component-wise transformations.  Separability is shown to impose constraints on certain locally invariant (scalar) functions of $x$, which are derived from local higher-order correlations of the data's velocity $\dot{x}$.  The data are separable if and only if they satisfy these constraints, and, if the constraints are satisfied, the sources can be explicitly constructed from the data. The method is illustrated by using it to separate two speech-like sounds recorded with a single microphone.
\end{abstract}

\section{Introduction}

Sensory devices often receive signals from multiple physical stimuli that evolve simultaneously but are unrelated to one another. In many of these situations, it is necessary to create separate representations of one or more of these stimuli by blindly processing the observed signals (i.e., by processing them without prior knowledge of the nature of the stimuli). In recent years, there has be considerable progress in the solution of this ``blind source separation" (BSS) problem for the special case in which the signals and source variables are linearly related. However, although nonlinear BSS is often performed effortlessly by humans, computational methods for doing this are quite limited~\cite{Jutten}.

Consider a time series of data $x(t)$, where $x$ is a multiplet of $N$ measurements ($x_k \mbox{ for } k = 1,2, \ldots ,N$). The usual objectives of nonlinear BSS are: 1) determine if these data are instantaneous mixtures of $N$ statistically independent source components $s(t)$
\begin{equation}
\label{mixture}
x(t) = f[s(t)] ,
\end{equation}
where $f$ is a possibly nonlinear, invertible $N \mbox{-component}$ mixing function; 2) if this is the case, compute the mixing function. In other words, the problem is to find a coordinate transformation $f^{-1}$ that transforms the observed data $x(t)$ from the measurement-defined coordinate system ($x$) on state space to a special source coordinate system ($s$) in which the components of the transformed data are statistically independent. Let $\rho_S(s)$ be the state space probability density function (PDF) in the source coordinate system, defined so that $\rho_S(s) ds$ is the fraction of total time that the source trajectory $s(t)$ is located within the volume element $ds$ at location $s$. In the usual formulation of the BSS problem, the source components are required to be statistically independent in the sense that their state space PDF is the product of the density functions of the individual components
\begin{equation}
\label{state space factorization}
\rho_S(s) = \prod_{k=1}^N{\rho_k(s_k)} .
\end{equation}
In every formulation of BSS, multiple solutions can be created by permutations and component-wise transformations of any one solution. However, it is well known that the criterion in (\ref{state space factorization}) is so weak that it suffers from a much worse non-uniqueness problem: namely, in this form of the BSS problem, multiple solutions can be created by transformations that mix the source variables (see~\cite{Hyvarinen-uniqueness} and references therein).

The issue of non-uniqueness can be circumvented by considering the data's trajectory in $(s,\dot{s}) \mbox{-space}$ ($\dot{s} = ds/dt$) instead of $s \mbox{-space}$ (i.e., state space). First, let $\rho_S(s,\dot{s})$ be the PDF in this space, defined so that $\rho_S(s,\dot{s}) ds d\dot{s}$ is the fraction of total time that the location and velocity of the source trajectory are within the volume element $ds d\dot{s}$ at location $(s,\dot{s})$. An earlier paper~\cite{Levin-bss-JAP} described a formulation of the BSS problem in which this PDF was required to be the product of the density functions of the individual components
\begin{equation}
\label{phase space factorization}
\rho_S(s,\dot{s}) = \prod_{k=1}^N{\rho_k(s_k,\dot{s}_k)} .
\end{equation}
Separability in $(s,\dot{s}) \mbox{-space}$ is a stronger requirement than separability in state space. To see this, note that (\ref{state space factorization}) can be recovered by integrating both sides of (\ref{phase space factorization}) over all velocities, but the latter equation cannot be deduced from the former one. In fact, it can be shown that (\ref{phase space factorization}) is strong enough to guarantee that the BSS problem in $(s,\dot{s}) \mbox{-space}$ has a unique solution, up to permutations and component-wise transformations~\cite{Levin-bss-JAP}.  Furthermore, this type of statistical independence has the virtue of being satisfied by almost all classical physical systems that are composed of non-interacting subsystems, which are the generators of most signals of interest.

The author previously demonstrated~\cite{Levin-bss-JAP} that the $(s,\dot{s}) \mbox{-space}$ PDF of a time series induces a Riemannian geometry on the state space, with the metric equal to the local second-order correlation matrix of the data's velocity. Nonlinear BSS can be performed by computing this metric in the $x$ coordinate system (i.e., by computing the second-order correlation of $\dot{x}$ at each point $x$), as well as its first and second derivatives with respect to $x$. However, although this is a mathematically correct and complete method of solving the nonlinear BSS problem, it suffers from a practical difficulty: namely, if the dimensionality of state space is high, a great deal of data is required to cover it densely enough in order to calculate these derivatives accurately. The current paper~\cite{Levin-ICA09} shows how to perform nonlinear BSS by computing higher-order local correlations of the data's velocity, instead of computing derivatives of its second-order correlation. This approach is advantageous because it requires much less data for an accurate computation. For example, in the synthetic speech separation experiment in Section III, the new method can separate two synthetic utterances recorded with a single microphone after minutes of observation, rather than the hours of observation required by the differential geometric method.

The method described in this paper differs significantly from the methods proposed by other investigators because it uses a criterion of statistical independence in $(s,\dot{s}) \mbox{-space}$, instead of state space. In addition, there are technical differences between the proposed method and conventional ones. First of all, the technique in this paper exploits statistical constraints on the data that are \textit{locally} defined in state space, in contrast to the usual criteria for statistical independence that are \textit{global} conditions on the data time series or its time derivatives~\cite{Lagrange}. Furthermore, unlike many other methods~\cite{Yang,Haykin}, the mixing function is derived in a constructive, deterministic, and non-parametric manner, without employing iterative algorithms, without using probabilistic learning methods, and without parameterizing it with a neural network architecture or other means. In addition, the proposed method can handle any differentiable mixing function, unlike some other techniques that only apply to a restricted class of mixing functions~\cite{Taleb}.

The next section describes how to separate two-dimensional data into two one-dimensional source variables. Section III illustrates the method by using it to separate two simultaneous speech-like sounds that are recorded with a single microphone.  The implications of this work are discussed in the last section. The appendix describes how the method can be generalized to separate data of arbitrary dimensionality into possibly multidimensional source variables.

\section{Method}

The BSS procedure, which is described in this section, is initiated by constructing scalar functions on the data space from combinations of local velocity correlations. The values of these scalars are invariant under any nonlinear transformations of coordinates on the data space. It is relatively easy to show that separability imposes necessary conditions on these scalar functions in the source coordinate system. Because of their scalarity, these conditions can readily be transferred to the measurement-defined coordinate system ($x$), where they can be tested with the data. If the data do not satisfy these necessary conditions, the data are simply not separable. If the data do satisfy these conditions, we show that there is only one possible source coordinate system, and it can be explicitly constructed. The data can then be transformed into this putative source coordinate system to see if their PDF and/or correlations factorize there. The data are separable if and only if this factorization occurs.
 
The first step is to construct local correlations of the data's velocity, such as
\begin{equation}
\label{C definition}
C_{kl \ldots}(x) = \, <(\dot{x}_k-\bar{\dot{x}}_k) (\dot{x}_l-\bar{\dot{x}}_l) \ldots>_{x} ,
\end{equation}
where $\bar{\dot{x}} = <\dot{x}>_x$, where the bracket denotes the time average over the trajectory's segments in a small neighborhood of $x$, where $1 \leq k, \, l \leq N$, and where ``$\ldots$" denotes possible additional indices on the left side and corresponding factors of $\dot{x}-\bar{\dot{x}}$ on the right side. The definition of the PDF implies that this velocity correlation is one of its moments
\begin{equation}
\label{PDF moment}
C_{kl \ldots}(x) = \frac {\int \rho(x,\dot{x}) (\dot{x}_k-\bar{\dot{x}}_k) (\dot{x}_l-\bar{\dot{x}}_l) \ldots d\dot{x}} {\int \rho(x,\dot{x}) d\dot{x}} ,
\end{equation}
where $\rho(x,\dot{x})$ is the PDF in the $x$ coordinate system. Incidentally, although (\ref{PDF moment}) is useful in a formal sense, in practical applications, all required correlation functions can be computed directly from local time averages of the data ((\ref{C definition})), without explicitly computing the data's PDF. Also, note that velocity ``correlations" with a single subscript vanish identically.

Next, let $M(x)$ be a local $N \, x \, N$ matrix, and use it to define $M \mbox{-transformed}$ velocity correlations
\begin{equation}
\label{I definition}
I_{kl \ldots}(x) = \sum_{1 \leq k', \, l', \ldots \leq N} M_{kk'}(x) M_{ll'}(x) \ldots C_{k'l' \ldots}(x) ,
\end{equation}
where ``$\ldots$" denotes possible additional indices of $I$ and $C$, as well as corresponding factors of $M(x)$. Because $C_{kl}(x)$ is positive definite at any point $x$, it is always possible to find an  $M(x)$ such that
\begin{equation}
\label{M definition 1}
I_{kl}(x) = \delta_{kl}
\end{equation}
\begin{equation}
\label{M definition 2}
\sum_{1 \leq m \leq N} I_{klmm}(x) = D_{kl}(x) , 
\end{equation}
where $D(x)$ is a diagonal $N \, x \, N$ matrix. Such an $M(x)$ can always be constructed from the product of three matrices: 1) a rotation that diagonalizes $C_{kl}(x)$, 2) a diagonal rescaling matrix that transforms this diagonalized correlation into the identity matrix, 3) another rotation that diagonalizes
\begin{displaymath}
\sum_{1 \leq m \leq N} C_{klmm}(x) ,
\end{displaymath} 
after the fourth-order correlation has been transformed by the first rotation and the rescaling matrix.   As long as the last-diagonalized matrix is not degenerate, $M(x)$ is unique, up to arbitrary \textit{local} permutations and reflections. In almost all realistic applications, the velocity correlations will be continuous functions of the state space coordinate $x$. Therefore, in any neighborhood of state space, there will always be a continuous solution for $M(x)$, and this solution is unique, up to arbitrary \textit{global} reflections and permutations.

In order to show that the $M \mbox{-transformed}$ velocity correlations (i.e., the $I_{klm \ldots}(x)$) transform like scalars, imagine constructing these quantities in some other coordinate system $x'$. An $M \mbox{-matrix}$ that satisfies (\ref{M definition 1}) and (\ref{M definition 2}) in the $x'$ coordinate system is given by
\begin{equation}
\label{M'}
M'_{kl}(x') = \sum_{1 \leq m \leq N} M_{km}(x) \frac{\partial x_m}{ \partial x'_l} (x') ,
\end{equation}
where $M$ is a matrix that satisfies (\ref{M definition 1}) and (\ref{M definition 2}) in the $x$ coordinate system. To prove this, substitute this equation into the definition of $I'_{kl \ldots}(x')$. Because velocity correlations transform as contravariant tensors, the partial derivative factors within $M'$ transform correlations from the $x'$ coordinate system to the $x$ coordinate system, leading to
\setlength\arraycolsep{2pt}
\begin{eqnarray*}
\label{I'}
I'_{kl \ldots}(x') & = & \sum_{1 \leq k', \, l', \ldots \leq N} M'_{kk'}(x') M'_{ll'}(x') \ldots C'_{k'l' \ldots}(x') \\
                   & = & \sum_{1 \leq k', \, l', \ldots \leq N} M_{kk'}(x) M_{ll'}(x) \ldots C_{k'l' \ldots}(x) \\
                   & = & I_{kl \ldots}(x) .
\end{eqnarray*}
Therefore, because $I_{kl}(x)$ and $I_{klmn}(x)$ satisfy (\ref{M definition 1}) and (\ref{M definition 2}), so do $I'_{kl}(x')$ and $I'_{klmn}(x')$, thereby proving that (\ref{M'}) is one of the solutions for $M'(x')$ in the $x'$ coordinate system. All other solutions for $M'(x')$ differ from this one by global reflections and permutations. Similar reasoning shows that, for any choice of $M'$ and $M$, each of the functions $I'_{kl \ldots}(x')$ equals the corresponding function $I_{kl \ldots}(x)$, up to possible global permutations and reflections.  In other words,
\begin{equation}
\label{I' = PI}
I'_{kl \ldots}(x') = \sum_{1 \leq k', \, l', \ldots \leq N} P_{kk'} P_{ll'} \ldots I_{k'l' \ldots}(x) ,
\end{equation}
where $P_{kk'}$ denotes an element of a product of permutation, reflection, and identity matrices. In other words, the functions $I_{kl \ldots}(x)$ transform as scalar functions on the state space, except for possible reflections and index permutations.

We now assume that the system is separable and derive some necessary conditions on these scalar functions in the source coordinate system ($s$). Because these separability conditions involve scalar functions, they can then be transferred to the measurement-defined coordinate system ($x$), where they can be tested with the data. In order to make the notation simple, it is assumed that $N = 2$ in the following. However, the appendix describes how the methodology can be generalized in order to separate higher-dimensional data into possibly multidimensional source variables.

Separability implies that there is a transformation $f^{-1}$ from the $x$ coordinate system to a source coordinate system ($s$) in which (\ref{phase space factorization}) is true. Because of (\ref{PDF moment}), the velocity correlation functions in the $s$ coordinate system are products of correlations of the independent sources
\begin{equation}
\label{C = C1 C2}
C_{S1 \ldots 2 \ldots}(s) = C_{S1 \ldots}(s_1) \, C_{S2 \ldots}(s_2) ,
\end{equation}
where $1 \ldots$ and $2 \ldots$ denote arbitrary numbers of indices equal to 1 and 2, respectively. It follows from this equation and from the vanishing of all velocity ``correlations" with one index that the source variable correlations $C_{Skl}(s)$ and 
\begin{displaymath}
\label{CSklmm}
\sum_{1 \leq m \leq N} C_{Sklmm}(x)
\end{displaymath}
are diagonal. Therefore, in the $s$ coordinate system, (\ref{M definition 1}) and (\ref{M definition 2}) are satisfied by a diagonal matrix $M_S(s)$ of the form 
\begin{equation}
\label{diagonal MS}
M_S(s) = \left( \begin{array}{cc}  
   M_{S1}(s_1) & 0 \\
   0            & M_{S2}(s_2)
   \end{array} \right) .
\end{equation}
It follows from (\ref{C = C1 C2}) and (\ref{diagonal MS}) that the scalar functions $I_{Skl \ldots}(s)$ with all subscripts $kl \ldots$ equal to 1 (2) must equal the corresponding functions derived for subsystem 1 (2), and these latter functions depend on $s_1$ ($s_2$) alone. Although these constraints were derived in the $s$ coordinate system, scalarity ((\ref{I' = PI})) implies that these separability conditions are true in all coordinate systems, except for possible permutations and reflections. Therefore, in the measurement-defined  coordinate system ($x$), the functions $I_{kl \ldots}(x)$ with all subscripts equal to 1 must be functions of either $s_1(x)$ \textit{or} $s_2(x)$. Likewise, the functions $I_{kl \ldots}(x)$ with all subscripts equal to 2 must be functions of the other source variable ($s_2(x)$ or $s_1(x)$, respectively).

This coordinate-system-independent consequence of separability can be used to perform nonlinear BSS in the following manner:
\begin{enumerate}
 \item Use (\ref{C definition}) to compute velocity correlations $C_{kl \ldots}(x)$ from the data $x(t)$.
 \item Use linear algebra to find a continuous matrix $M(x)$ that satisfies (\ref{M definition 1}) and (\ref{M definition 2}).
 \item Use (\ref{I definition}) to compute the functions $I_{kl \ldots}(x)$.
 \item Plot the values of the triplets
\setlength\arraycolsep{2pt}
\begin{equation}
\label{IA multiplet}
I_A(x) = \{I_{111}(x), \, I_{1111}(x), \, I_{11111}(x) \}
\end{equation}
\begin{equation}
\label{IB multiplet}
I_B(x) = \{I_{222}(x), \, I_{2222}(x), \, I_{22222}(x) \} 
\end{equation}
as $x$ varies over the measurement-defined coordinate system.
 \item If the plotted values of $I_A$ and/or $I_B$ \textit{do not} lie in one-dimensional subspaces within the three-dimensional space of the plots,  $I_A(x)$ and/or $I_B(x)$ cannot be functions of single source components ($s_1(x)$ or $s_2(x)$) as required by separability, and the data are not separable.
 \item If the plotted values of both $I_A$ and $I_B$ \textit{do} lie on one-dimensional manifolds, define one-dimensional coordinates ($\sigma_A$ and $\sigma_B$, respectively) on those subspaces. Then, compute the function $\sigma(x) = (\sigma_A(x),\sigma_B(x))$ that maps each coordinate $x$ onto the value of $\sigma$, which parameterizes the point $(I_A(x), I_B(x))$. Notice that, because of the Takens' embedding theorem~\cite{Sauer}, $x$ is invertibly related to the six components of $I_A(x)$ and $I_B(x)$, and, therefore, it is invertibly related to $\sigma$.
 \item Transform the PDF (or correlations) of the measurements from the $x$ coordinate system to the $\sigma$ coordinate system. The data are separable if and only if the PDF factorizes (the correlations factorize) in the $\sigma$ coordinate system.
\end{enumerate}

The last statement can be understood in the following manner. As shown above, separability implies that $I_A(x)$ must be a function of a single source variable ($s_1(x)$ or $s_2(x)$), and the Takens theorem implies that this function is invertible.  Because $I_A$ is also an invertible function of $\sigma_A$, it follows that $\sigma_A$ must be invertibly related to one of the source variables, and, in a similar manner, $\sigma_B$ must be invertibly related to the other source variable. Thus, separability implies that $\sigma_A$ and $\sigma_B$ are themselves source variables. It follows that the data are separable if and only if the PDF factorizes in the $\sigma$ coordinate system.

Although the above-described procedure will perform BSS for any mixing function, it is interesting to consider the special case in which source variables exist that are linearly related to the measurements; namely,
\begin{equation}
\label{s = Wx}
s_k = \sum_{1 \leq l \leq 2} W_{kl} x_l ,
\end{equation}
where $W$ is a constant $2 \, x \, 2$ matrix. In general, the above BSS procedure will construct source variables $\sigma$ that are related to these ``linear" source variables by
\setlength\arraycolsep{2pt}
\begin{eqnarray}
\label{sigma = g(s)}
\sigma_1(x) & = & g_1(s_1(x)) = g_1(W_{11} x_1 + W_{12} x_2) \\
\sigma_2(x) & = & g_2(s_2(x)) = g_2(W_{21} x_1 + W_{22} x_2) ,
\end{eqnarray}
where $g_1$ and $g_2$ are some invertible nonlinear transformations determined by the choice of the $\sigma_1$ and $\sigma_2$ coordinates, respectively. Therefore, at each point $x$ the partial derivatives
\begin{eqnarray*}
\partial \sigma_1 / \partial x_k \\ 
\partial \sigma_2 / \partial x_k
\end{eqnarray*}
will be proportional to constant (i.e., $x \mbox{-independent}$) vectors (denoted by $U_{1k}$ and $U_{2k}$), which are themselves proportional to the first and second rows of $W$, respectively. Furthermore, these vectors can be used to construct other linearly-related source variables
\begin{equation}
\label{sHat = Ux}
\hat{s}_k = \sum_{1 \leq l \leq 2} U_{kl} x_l
\end{equation}
that are just rescaled versions of the ones in (\ref{s = Wx}). Consequently, given the source variables $\sigma(x)$ produced by the BSS procedure, the following process can be used to determine whether these can be transformed into source variables that are linearly related to the measurements: 1) compute the above-mentioned partial derivatives and determine if each is proportional to an $x \mbox{-independent}$ vector; 2) if the partial derivatives do not satisfy this condition, there are no linearly-related source variables; 3) if the partial derivative do  satisfy condition 1, transform the data into the $\hat{s}$ coordinate system in order to see if the data's PDF factorizes there. There are linearly-related source variables if and only if this factorization occurs.

\begin{figure*}[!t]
\centerline{\subfigure[]{\includegraphics[width=2.2in]{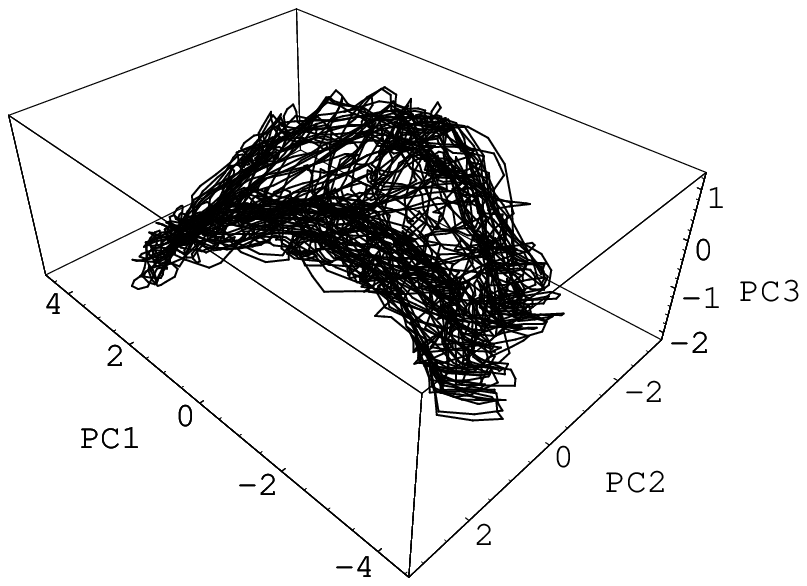}%
\label{figure1a}}
\hfil
\subfigure[]{\includegraphics[height=2.2in]{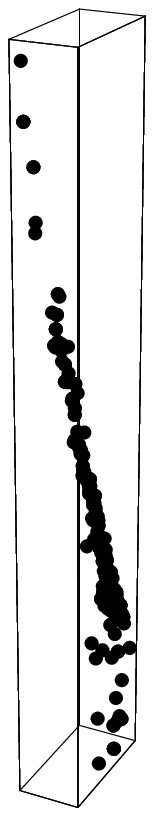}%
\label{figure1b}}
\hfil
\subfigure[]{\includegraphics[height=2.2in]{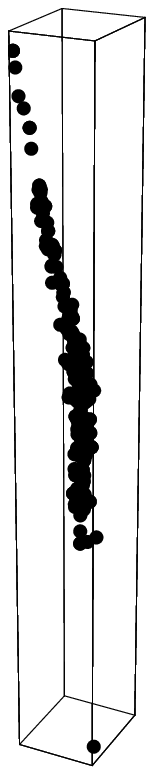}%
\label{figure1c}}}
\caption{(a) The first three principal components of log filterbank outputs of a typical short recording of two simultaneous speech-like sounds. (b) The distribution of the values of $I_A(x)$ ((\ref{IA multiplet})), as $x$ varied over the approximately two-dimensional manifold in (a). (c) The distribution of the values of $I_B(x)$ ((\ref{IB multiplet})), as $x$ varied over the approximately two-dimensional manifold in (a).}
\label{figure1}
\end{figure*}

\section{Numerical Example: Separating Two Speech-Like Sounds Recorded with a Single Microphone}

\begin{figure*}[!t]
\centerline{\subfigure[]{\includegraphics[width=2.2in]{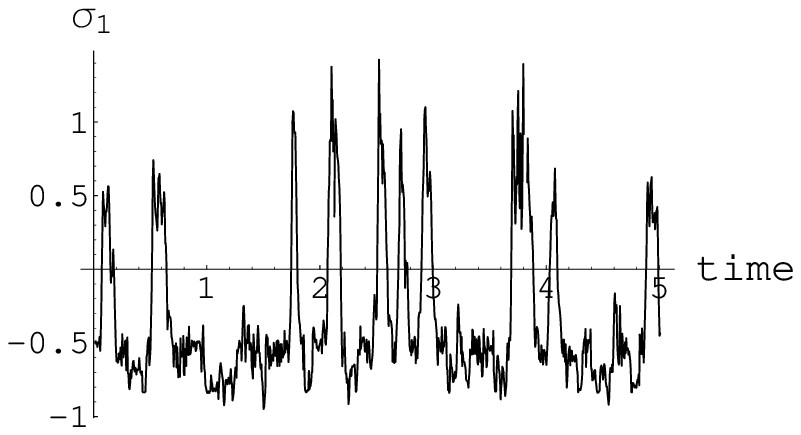}%
\label{figure2a}}
\hfil
\subfigure[]{\includegraphics[width=2.2in]{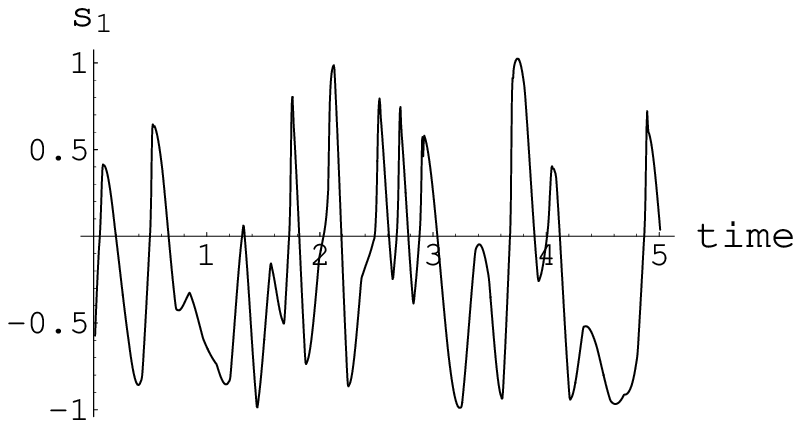}%
\label{figure2b}}
\hfil
\subfigure[]{\includegraphics[width=2.2in]{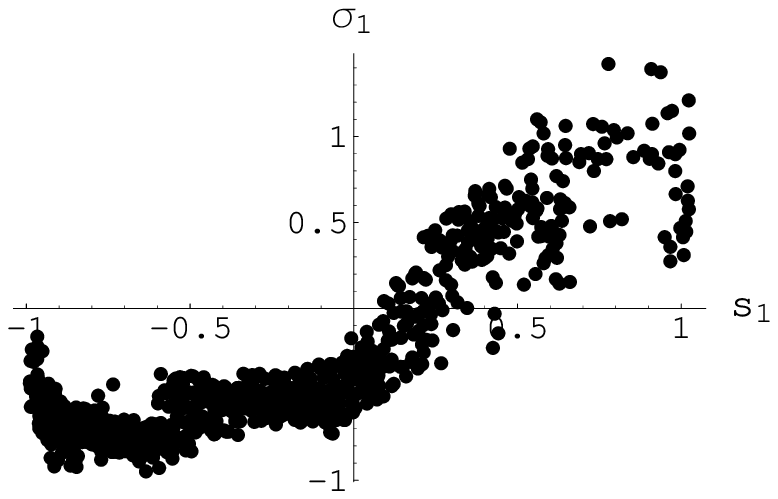}%
\label{figure2c}}}
\caption{(a) The time dependence of one of the source variables, blindly computed from a typical five-second segment of the data's trajectory $x(t)$, by finding the coordinates of $I_A[x(t)]$ on the one-dimensional manifold in Fig.~1b. (b) The state variable time series originally used to generate one of the speech-like sounds during the five-second recording analyzed in (a). (c) The scatter plot of the pairs of source and state variable values in (a) and (b).}
\label{figure2}
\end{figure*}

\begin{figure*}[!t]
\centerline{\subfigure[]{\includegraphics[width=2.2in]{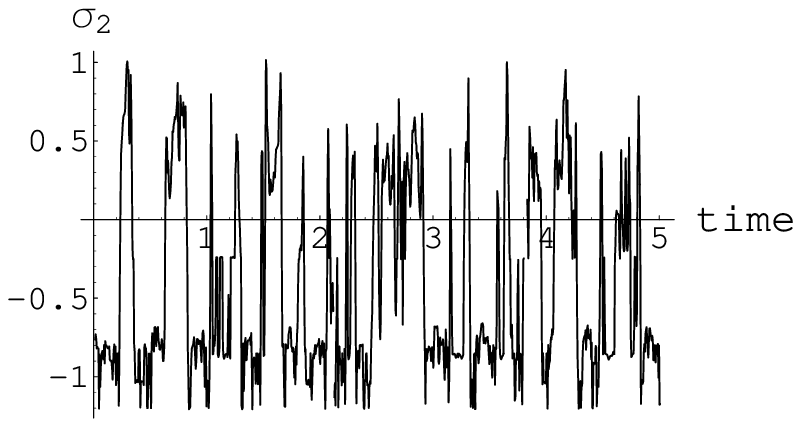}%
\label{figure3a}}
\hfil
\subfigure[]{\includegraphics[width=2.2in]{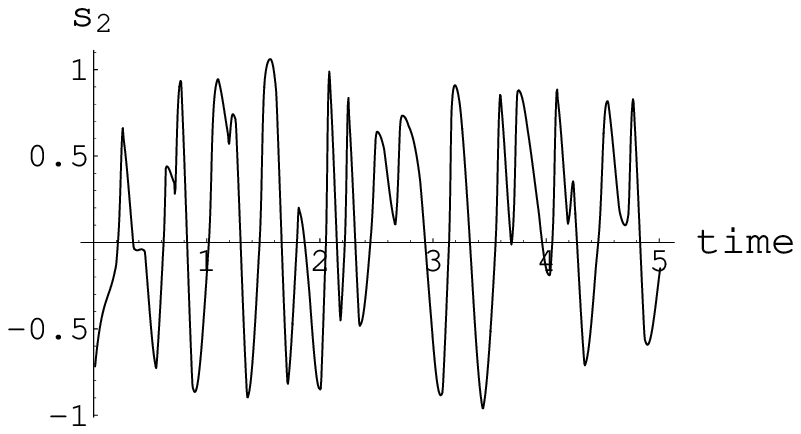}%
\label{figure3b}}
\hfil
\subfigure[]{\includegraphics[width=2.2in]{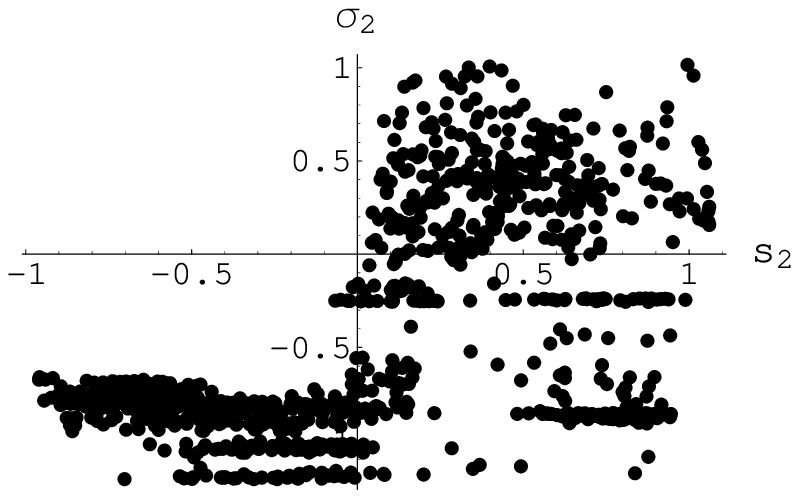}%
\label{figure3c}}}
\caption{(a) The time dependence of one of the source variables, blindly computed from a typical five-second segment of the data's trajectory $x(t)$, by finding the coordinates of $I_B[x(t)]$ on the one-dimensional manifold in Fig.~1c. (b) The state variable time series originally used to generate one of the speech-like sounds during the five-second recording analyzed in (a). (c) The scatter plot of the pairs of source and state variable values in (a) and (b).}
\label{figure3}
\end{figure*} 

This section describes a numerical experiment in which two speech-like sounds were synthesized and then summed, as if they were simultaneously recorded with a single microphone. Each sound simulated an ``utterance" of a vocal tract resembling a human vocal tract, except that: 1) it had one degree of freedom, instead of the 3-5 degrees of freedom of the human vocal tract; 2) its impulse response was characterized by one pole pair, instead of the 4-6 pole pairs characteristic of the human vocal tract. The methodology of Section II was blindly applied to a time series of two features extracted from the synthetic recording, in order to recover the time dependence of the state variable of each vocal tract (up to an unknown transformation on each voice's state space). BSS was performed with only 16 minutes of data, instead of the hours of data required to separate similar sounds using a differential geometric method~\cite{Levin-bss-JAP}.

Each speaker was simulated by having a simulated glottis drive a simulated resonant cavity that represented the vocal tract. The glottal waveform of the each ``voice" was a series of spikes separated by a pitch interval (100 Hz and 160 Hz). The impulse response of each ``vocal tract" was taken to be a characteristic damped sinusoid, whose amplitude, resonant frequency, and damping were linear functions of a single state variable. For each voice, a 16 minute utterance was produced by convolving its glottal waveform with the impulse response of its vocal tract, which was a function of a slowly-varying state variable. The state variable time series of each voice was synthesized by smoothly interpolating among successive states. The latter were chosen at 100-120 msec intervals so that the state variable time series of the two voices were statistically independent of each other. The resulting utterances had energies differing by 2.4 dB, and they were summed and sampled at 16 kHz with 16-bit depth.  Then, this ``recorded" waveform was pre-emphasized and subjected to a short-term Fourier transform (using frames with 25 msec length and 5 msec spacing).  The log energies of a bank of 12 mel-frequency filters between 0-8000 Hz were computed for each frame, and these were then averaged over pairs of consecutive frames.  These log filterbank outputs were nonlinear functions of the two vocal tract state variables.

In order to blindly analyze these data, we first determined if any data components were redundant in the sense that they were simply functions of other components.  Fig.~\ref{figure1}a shows the first three principal components of the log filterbank outputs during a typical short recording of the simultaneous utterances.  Inspection showed that these data lay on a curved two-dimensional surface within the ambient 12-D space, making it apparent that they were produced by a ``hidden" dynamical system with two degrees of freedom.  The redundant components were eliminated by using dimensional reduction (principal components analysis in small overlapping neighborhoods of the data) to establish a coordinate system $x$ on this surface and to find $x(t)$, the trajectory of the recorded sound in that coordinate system.  Next, the BSS procedure in Section II was used to determine if $x(t)$ was a nonlinear mixture of two source variables that were statistically independent of one another.  Following steps 1-4 of the BSS procedure, $x(t)$ of the entire recording was used to compute invariants $I_{kl \ldots}(x)$ with up to five indices, and the related functions $I_A(x)$ and $I_B(x)$ were plotted, as illustrated in Figs.~\ref{figure1}b-c. It was evident that the plotted values of both $I_A$ and $I_B$ lay in or close to one-dimensional subspaces. Following step 6 of the BSS procedure, a dimensional reduction procedure~\cite{Roweis} was used to define coordinates ($\sigma_A$ and $\sigma_B$) on these one-dimensional manifolds, and $\sigma(x) = (\sigma_A(x),\sigma_B(x))$ was computed. If the data were separable, $\sigma$ must be a set of source variables, and $\sigma[x(t)]$ must describe the evolution of the underlying vocal tract states (up to invertible component-wise transformations). As illustrated in Figs.~\ref{figure2}a-b and Figs.~\ref{figure3}a-b, the time courses of the putative source variables ($\sigma_A[x(t)],\sigma_B[x(t)]$) did resemble distorted versions of the state variable time series that were originally used to generate the voices' utterances. The scatter plots in Fig.~\ref{figure2}c and Fig.~\ref{figure3}c show that, in each case, the recovered source variable and the corresponding state variable were related by a nonlinear transformation that was nearly monotonic, except for the effects of noise due to the limited number of data samples. Thus, starting with a single-microphone recording, the BSS procedure was able to extract the information encoded in the time series of each speaker's state variable. The time course of the analogous multidimensional state variable of the human vocal tract contains the speech content of each utterance. This indicates that the BSS procedure is capable of recovering the speech content of superposed utterances, without recovering their original waveforms.

\section{Discussion}

In a previous paper~\cite{Levin-bss-JAP}, the nonlinear BSS problem was reformulated in (state, state velocity)-space, instead of state space as in conventional formulations. This approach is attractive because: 1) the reformulated BSS problem has a unique solution in the following sense: either the data are inseparable, or they can be separated by a mixing function that is unique (up to permutations and transformations of independent source variables); 2) statistical independence in (state, state velocity)-space is manifested by almost all classical physical systems that are composed of non-interacting subsystems. This paper~\cite{Levin-ICA09} shows how a general solution of this problem can be constructed in a deterministic manner, which avoids the difficulties of the iterative, probabilistic, and parametric BSS techniques proposed by other investigators. Furthermore, an accurate computation can be performed with far less data than that required by the differential geometric solution, previously proposed by the author~\cite{Levin-bss-JAP}.

The BSS procedure in Section II shows how to compute $(\sigma_A[x(t)], \sigma_B[x(t)])$, the trajectory of each independent subsystem in a specific coordinate system on that subsystem's state space. In many practical applications, a pattern recognition ``engine" has been trained to recognize the meaning of trajectories of one subsystem (e.g., ``$A$") in another coordinate system (e.g., $s_A$) on that subsystem's state space. In order to use this information, it is necessary to know the transformation to this particular coordinate system ($\sigma_A \rightarrow s_A$). For example, subsystem $A$ may be the vocal tract of speaker $A$, and subsystem $B$ may be a noise generator of some sort. In this example, we may have trained an automatic speech recognition (ASR) engine on the quiet speech of speaker $A$ (or, equivalently, on the quiet speech of another speaker who mimics $A$ in the sense that their state space trajectories are related by an invertible transformation when they speak the same utterances). In order to recognize the speaker's utterances in the presence of $B$, we must know the transformation from the vocal tract coordinates recovered by BSS ($\sigma_A$) to the coordinates used to train the ASR engine ($s_A$). This mapping can be determined by using the training data to compute more than $2d_A$ invariants (like those in (\ref{I definition})) as functions of $s_A$. These must equal the invariants of one of the subsystems identified by the BSS procedure, up to a global permutation and/or reflection ((\ref{I' = PI})). This global transformation can be determined by permuting and reflecting the distribution of invariants produced by the training data, until it matches the distribution of invariants of one of the subsystems produced by the BSS procedure. Then, the mapping $\sigma_A \rightarrow s_A$ can be determined by finding paired values of $\sigma_A$ and $s_A$ that correspond to the same invariant values within these matching distributions. This type of analysis of human speech data is currently underway.

\appendix[Separating data of any dimensionality \\ into possibly multidimensional sources]

The procedure in Section II is capable of separating two-dimensional data into one-dimensional source variables. This appendix describes the solution of the more general nonlinear BSS problem in which data of any dimensionality may be separated into possibly multidimensional source variables, each of which is statistically independent of the others but each of which may contain statistically dependent components. This is sometimes called multidimensional independent component analysis, subspace independent component analysis, or independent subspace analysis~\cite{Cardoso,Nishimori}.

Separability implies that there is a transformation $f^{-1}$ from the $x$ coordinate system to a source coordinate system ($s = (s_A,s_B)$) in which 
\begin{equation}
\label{rho = rhoA rhoB}
\rho_S(s,\dot{s})=\rho_A(s_A,\dot{s}_A) \rho_B(s_B,\dot{s}_B) ,
\end{equation}
where $s_A$ is a possibly multidimensional source variable with $d_A$ components and $s_B$ is a possibly multidimensional source variable with $d_B = N - d_A$ components. Because of (\ref{PDF moment}), the velocity correlation functions in the $s$ coordinate system are products of correlations of independent sources
\begin{equation}
\label{C = CA CB}
C_{Sa \ldots b \ldots}(s) = \, C_{Sa \ldots}(s_A) \, C_{Sb \ldots}(s_B) ,
\end{equation}
where $a \ldots$ and $b \ldots$ denote arbitrary series of indices in the ranges $1 \leq a \leq d_A$ and $d_A+1 \leq b \leq N$, respectively. It follows from this equation and from the vanishing of all velocity ``correlations" with one index that the source variable correlations $C_{Skl}(s)$ and 
\begin{displaymath}
\label{CSklmm 2}
\sum_{1 \leq m \leq N} C_{Sklmm}(x)
\end{displaymath}
have block-diagonal forms with $d_A \, x \, d_A$ and $d_B \, x \, d_B$ upper and lower blocks, respectively. Consequently, in the $s$ coordinate system, (\ref{M definition 1}) and (\ref{M definition 2}) are satisfied by a block-diagonal matrix $M_S(s)$ of the form 
\begin{equation}
\label{block-diagonal MS}
M_S(s) = \left( \begin{array}{cc}  
   M_{SA}(s_A) & 0 \\
   0            & M_{SB}(s_B)
   \end{array} \right) ,
\end{equation}
where $M_{SA}$ and $M_{SB}$ are matrices that satisfy (\ref{M definition 1}) and (\ref{M definition 2}) for the $A$ and $B$ subsystems, respectively. In order to prove that (\ref{block-diagonal MS}) satisfies (\ref{M definition 1}), substitute it into the definition of $I_{Skl}$, and note that each block of $M_S$ is defined to transform the corresponding block of $C_{Skl}$ into an identity matrix. In order to prove that (\ref{block-diagonal MS}) satisfies (\ref{M definition 2}), substitute it into the definition of
\begin{equation}
\label{ISklmm}
\sum_{1 \leq m \leq N} I_{Sklmm} .
\end{equation}
Then, note that: 1) when $k$ and $l$ belong to different blocks, each term in this sum vanishes because it factorizes into a product of a one-index correlation and a three-index correlation; 2) when $k$ and $l$ belong to the same block and are unequal, each term with $m$ in the other block contains a factor equal to $I_{Skl}$, which vanishes, as proved above; 3) when $k$ and $l$ belong to the same block and are unequal, the sum over $m$ in the same block vanishes, because each block of $M_S$ is defined to satisfy (\ref{M definition 2}) for the corresponding subsystem.  

It follows from (\ref{I definition}), (\ref{C = CA CB}), and (\ref{block-diagonal MS}) that the scalar functions $I_{Skl \ldots}(s)$ with all subscripts $kl \ldots$ in the range $1 \leq k, \, l \leq d_A$ (or in the range $d_A+1 \leq k, \, l \leq N$) must depend on $s_A$ (or $s_B$) alone. Although these constraints were derived in the $s$ coordinate system, scalarity ((\ref{I' = PI})) implies that these separability conditions are true in all coordinate systems, except for possible permutations. Therefore, in the measurement-defined  coordinate system ($x$), it must be possible to partition the indices of $x$ ($k = 1,2, \ldots ,N$) into $A$ and $B$ groups (containing $d_A$ ``$A$" indices and $d_B$ ``$B$" indices, respectively) so that the functions $I_{kl \ldots}(x)$ with all subscripts in the $A$ (or $B$) group are functions of $s_A(x)$ (or $s_B(x)$) alone.

This coordinate-system-independent consequence of separability can be used to perform nonlinear BSS in the following manner:
\begin{enumerate}
 \item Use (\ref{C definition}) to compute velocity correlations $C_{kl \ldots}(x)$ from the data $x(t)$.
 \item Use linear algebra to find a continuous matrix $M(x)$ that satisfies (\ref{M definition 1}) and (\ref{M definition 2}).
 \item Use (\ref{I definition}) to compute the functions $I_{kl \ldots}(x)$.
 \item Consider each choice of an integer $d_A$ in the range $1 \leq d_A < N$, and consider each way of partitioning the data indices ($k = 1,2, \ldots ,N$) into $A$ and $B$ groups (containing $d_A$ ``$A$" indices and $d_B = N - d_A$ ``$B$" indices, respectively). For each of these choices, let $I_A(x)$ ($I_B(x)$) be any set of more than $2d_A$ ($2d_B$) of the functions $I_{kl \ldots}(x)$ for which all subscripts belong to the $A$ ($B$) group, and plot the values of $I_A(x)$ and $I_B(x)$ as $x$ varies over the measurement-defined coordinate system.
 \item Suppose that, for all of the choices in step 4, the plotted values of $I_A$ and/or $I_B$ \textit{do not} lie in $d_A \mbox{-dimensional}$ ($d_B \mbox{-dimensional}$) subspaces within the higher-dimensional space of the plots. Then, there is no way that $I_A(x)$ and $I_B(x)$ can be functions of single source variables ($s_A(x)$ and $s_B(x)$) as required by separability, and the data are not separable.
 \item Suppose that, for one or more of the choices in step 4, the plotted values of both $I_A$ and $I_B$ \textit{do} lie in $d_A \mbox{-dimensional}$ and $d_B \mbox{-dimensional}$ manifolds, respectively. In that case, define $d_A \mbox{-dimensional}$ ($d_B \mbox{-dimensional}$) coordinates $\sigma_A$ ($\sigma_B$) on those subspaces. Then, compute the function $\sigma(x) = (\sigma_A(x), \sigma_B(x))$ that maps each coordinate $x$ onto the value of $\sigma$, which parameterizes the point $(I_A(x), I_B(x))$. Notice that, because of the Takens' embedding theorem~\cite{Sauer}, $x$ is invertibly related to the $2N+2$ (or more) components of $I_A(x)$ and $I_B(x)$, and, therefore, it is invertibly related to $\sigma$.
 \item Transform the PDF (or correlations) of the measurements from the $x$ coordinate system to the $\sigma$ coordinate system. The data are separable, and $\sigma_A$ and $\sigma_B$ are source variables if and only if the PDF factorizes (the correlations factorize) in a $\sigma$ coordinate system created in this way. 
\end{enumerate}

The last statement can be understood in the following manner. As shown above, separability implies that, for some choice of $d_A$ and index partitioning, $I_A(x)$ must be a function of $s_A(x)$, and the Takens theorem implies that this function is invertible.  Because $I_A$ is also an invertible function of $\sigma_A$, it follows that $\sigma_A$ must be invertibly related to $s_A$, and, in a similar manner, $\sigma_B$ must be invertibly related to $s_B$. Thus, separability implies that $\sigma_A$ and $\sigma_B$ are themselves source variables and, therefore, the PDF factorizes in the $\sigma$ coordinate system. Finally, note that, if the data are separable, the same procedure can then be used to determine if each multicomponent source variable ($\sigma_A$ or $\sigma_B$) can be further separated into lower-dimensional source variables.

The above-described procedure will perform BSS for any linear or nonlinear mixing. However, a few comments should be made about the special case in which source variables exist that are linearly related to the measurements; namely,
\begin{equation}
\label{sA = WAx}
s_{Aa} = \sum_{1 \leq k \leq N} W_{Aak} x_k
\end{equation}
\begin{equation}
\label{sB = WBx}
s_{Bb} = \sum_{1 \leq k \leq N} W_{Bbk} x_k ,
\end{equation}
where $1 \leq a \leq d_A$, where $1 \leq b \leq d_B$, and where $W_A$ and $W_B$ are constant $d_A \, x N$ and $d_B \, x N$ matrices, respectively. In general, the above BSS procedure will construct source variables $\sigma$ that are related to these ``linear" source variables by
\setlength\arraycolsep{2pt}
\begin{eqnarray}
\label{sigmaAB = g(sAB)}
\sigma_A & = & g_A(s_A) \\
\sigma_B & = & g_B(s_B) ,
\end{eqnarray}
where $g_A$ and $g_B$ are some nonlinear transformations (with $d_A$ and $d_B$ components, respectively), determined by the choice of the $\sigma_A$ and $\sigma_B$ coordinates, respectively. Therefore, at each point $x$, the sets of partial derivatives
\begin{eqnarray*} 
\partial \sigma_{Aa} / \partial x_k \mbox{ for } 1 \leq a \leq d_A \\
\partial \sigma_{Bb} / \partial x_k \mbox{ for } 1 \leq b \leq d_B
\end{eqnarray*} 
will be lie in the subspace spanned by the rows of $W_A$ and $W_B$, respectively. Let $U_{Aak}$ and $U_{Bbk}$ (for $1 \leq a \leq d_A$ and $1 \leq b \leq d_B$) denote any sets of constant vectors that span these two subspaces. Then, another set of linearly-related source variables is given by
\setlength\arraycolsep{2pt}
\begin{eqnarray}
\label{sHatAB = UABx}
\hat{s}_{Aa} & = & \sum_{1 \leq k \leq N} U_{Aak} x_k \\
\hat{s}_{Bb} & = & \sum_{1 \leq k \leq N} U_{Bbk} x_k ,
\end{eqnarray}
which are just linear combinations of the ones in (\ref{sA = WAx}) and (\ref{sB = WBx}), respectively. Consequently, given the source variables $\sigma(x)$ produced by the BSS procedure, the following process can be used to determine whether these can be transformed into source variables that are linearly related to the measurements: 1) compute the above-mentioned sets of partial derivatives and determine if each set is spanned by the appropriate number of $x \mbox{-independent}$ vectors; 2) if one or both sets of partial derivatives do not satisfy this condition, there are no linearly-related source variables; 3) if both sets of partial derivative do satisfy condition 1, transform the data into the $\hat{s}$ coordinate system in order to see if the data's PDF factorizes there. There are linearly-related source variables if and only if this factorization occurs.

%\section{References}

%\bibliographystyle{splncs}

\end{document}